\documentclass[sigconf]{acmart}
\usepackage{multirow}
\usepackage{booktabs}
\usepackage{amsthm}
\usepackage{graphicx}
\usepackage{amsmath}
\usepackage{color,xcolor}
\usepackage{hyperref}

\definecolor{mygreen}{RGB}{34,200,34}
\definecolor{myyellow}{RGB}{255,200,0}
\usepackage[linesnumbered,ruled,vlined]{algorithm2e}

\AtBeginDocument{%
  \providecommand\BibTeX{{%
    \normalfont B\kern-0.5em{\scshape i\kern-0.25em b}\kern-0.8em\TeX}}}

\setcopyright{acmcopyright}
\copyrightyear{2024}
\acmYear{2024}
\setcopyright{acmlicensed}\acmConference[ICMR '24]{Proceedings of the 2024 International Conference on Multimedia Retrieval}{June 10--14, 2024}{Phuket, Thailand}
\acmBooktitle{Proceedings of the 2024 International Conference on Multimedia Retrieval (ICMR '24), June 10--14, 2024, Phuket, Thailand}
\acmDOI{10.1145/3652583.3658038}
\acmISBN{979-8-4007-0619-6/24/06}

\begin{document}

\title{UBiSS: A Unified Framework for \\Bimodal Semantic Summarization of Videos}


\author{Yuting Mei}
\email{meiyuting1004@ruc.edu.cn}
\orcid{0009-0009-6036-8687}
\affiliation{
  \institution{Renmin University of China}
  \city{Beijing}
  \country{China}
}

\author{Linli Yao}
\email{linliyao@stu.pku.edu.cn}
\orcid{0000-0002-9809-8864}
\affiliation{
  \institution{Peking University}
  \city{Beijing}
  \country{China}
}

\author{Qin Jin}
\email{qjin@ruc.edu.cn}
\authornote{corresponding author}
\orcid{0000-0001-6486-6020}
\affiliation{
  \institution{Renmin University of China}
  \city{Beijing}
  \country{China}
}


\begin{abstract}

With the surge in the amount of video data, video summarization techniques, including visual-modal(VM) and textual-modal(TM) summarization, are attracting more and more attention. 
However, unimodal summarization inevitably loses the rich semantics of the video. 
In this paper, we focus on a more comprehensive video summarization task named {\textcolor{black}{Bi}modal \textcolor{black}{S}emantic \textcolor{black}{S}ummarization of Videos}(\textbf{\textcolor{black}{BiSSV}}). 
Specifically, we first construct a large-scale dataset, \textbf{BIDS}, in(video, VM-Summary, TM-Summary) triplet format. 
Unlike traditional processing methods, our construction procedure contains a VM-Summary extraction algorithm aiming to preserve the most salient content within long videos. 
Based on BIDS, we propose a Unified framework \textbf{\textit{UBiSS}} for the BiSSV task, which models the saliency information in the video and generates a TM-summary and VM-summary simultaneously. 
We further optimize our model with a list-wise ranking-based objective to improve its capacity to capture highlights. 
Lastly, we propose a metric, $NDCG_{MS}$, to provide a joint evaluation of the bimodal summary. 
Experiments show that our unified framework achieves better performance than multi-stage summarization pipelines. 
Code and data are available at \textcolor{blue}{\textit{\url{https://github.com/MeiYutingg/UBiSS}.}}

\end{abstract}


\maketitle

\section{Introduction}
\label{sec: introduction}

Video is becoming a critical information source on the Internet with a significant surge in volume\footnote{https://www.wyzowl.com/youtube-stats/}. 
Video summarization has emerged as a crucial technology for efficiently browsing, retrieving, and recommending videos. Based on modality, previous approaches could be divided into two categories: visual-modal summarization and textual-modal summarization of videos.

\begin{figure}[t]
\centering
\includegraphics[width=0.95\linewidth]{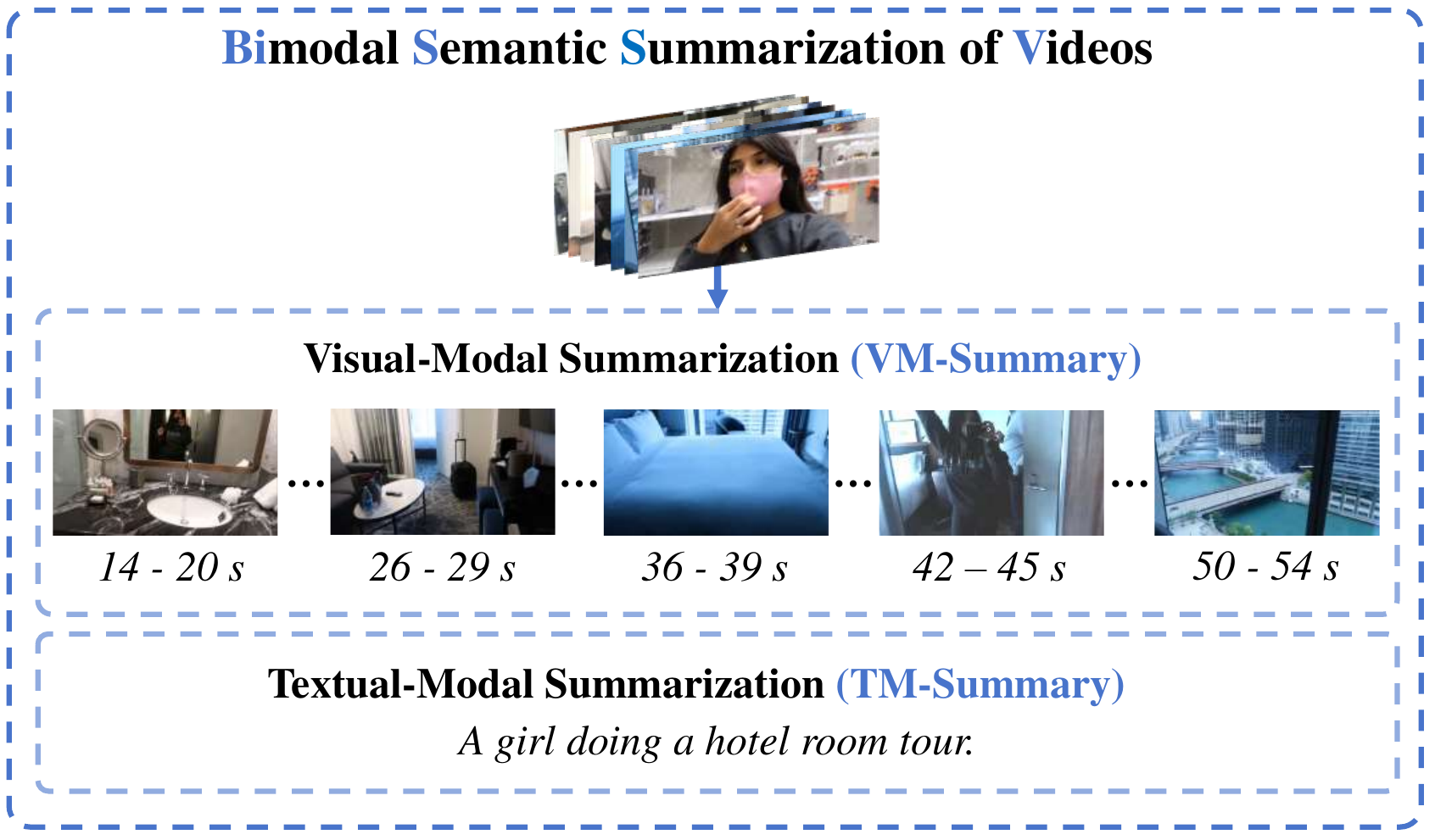}
\vspace{-10pt}
\caption{Illustration of the Bimodal Semantic Summarization of Videos (BiSSV) task, which generates video summaries in both textual-modality and visual-modality.} 
\vspace{-10pt}
\label{fig1}
\end{figure}

As the video content is inherently multimodal, unimodal summarization approaches, whether focusing on selecting informative clips~\cite{pang2023contrastive,narasimhan2022tl,apostolidis2021combining,8667390,ji2020deep,wang2019stacked,fajtl2019summarizing,lebron2018video,zhang2016video} or generating text descriptions~\cite{xu2023mplug,yang2023vid2seq,lin2022swinbert,wang2022git,venugopalan2015sequence}, inevitably sacrifice semantic richness in their pursuit of brevity. 
As shown in Figure ~\ref{fig1}, most textual-modal summarization approaches tend to generate a global summary description of the video content (\textit{e.g., "A girl doing a hotel room tour"}), while many visual details are lost. 
On the other hand, visual-modal summarization approaches provide highlighted and detailed visual information (\textit{e.g., a viewer can tell what the hotel room is like browsing through the summary clips}). However, since summary clips are unfolded sequentially, they cannot immediately convey the global core semantic content of the video. 

To strike a balance between offering a quick global overview and preserving the rich visual semantics of the original video, we focus on the task of \textbf{\textcolor{blue}{Bi}modal \textcolor{blue}{S}emantic \textcolor{blue}{S}ummarization of \textcolor{blue}{V}ideos} (\textbf{\textcolor{blue}{BiSSV}}) in this work. Our goal is to provide both \textit{textual-modal summary} (\textit{TM-Summary}) and \textit{visual-modal summary} (\textit{VM-Summary}) for videos. 
Creating such a bimodal summary is not trivial since it requires a thorough comprehension of both modalities and their relationship. An early attempt~\cite{chen2017video} fails to guarantee the correlation between bimodal summaries due to a lack of paired annotations. More recently, Lin et al.~\cite{lin2023videoxum} construct the first video-to-video+text dataset VideoXum, which enables simultaneous training for BiSSV. However, the TM-Summary in VideoXum comprises concatenated dense captions~\cite{krishna2017dense} with an average length of 49.9 words, losing the merit of conciseness of textual-modal summaries.

We consider that an ideal BiSSV dataset should follow specific criteria: For visual-modal summarization, the VM-Summary should preserve the main content of the video while following a length constraint to ensure conciseness. For textual-modal summarization, the TM-Summary should provide a highly abstract overview rather than present detailed descriptions of the video in chronological order. 
Since there are no existing datasets that meet these criteria, we build a \textbf{B}imodal V\textbf{ID}eo \textbf{S}ummarization dataset (\textbf{BIDS}) based on QVHighlights~\cite{lei2021detecting} through a three-step construction process, including data merging, VM-Summary extraction, and data cleaning. We design our VM-Summary extraction algorithm with a focus on preserving the most salient information within long segments, ensuring that the VM-Summary captures the gist effectively.

Furthermore, we propose a Unified framework for Bimodal semantic summarization of videos, namely \textcolor{black}{\textbf{{UBiSS}}}, which leverages cross-modal interaction for joint training and bimodal summarization. 
Although it may seem logical to consider BiSSV as two distinct subtasks (visual-model and textual-modal summarization), our experiments demonstrate that such a separated pipeline-type solution fails to recognize the natural connection between the two modalities. 
In contrast, our end-to-end approach, UBiSS, proves to be a more optimal solution. 
Furthermore, we observe that the traditional regression loss function may hinder the model from learning saliency trends. Therefore, we adopt a ranking-based list-wise optimization objective~\cite{Pobrotyn2021NeuralNDCG} to better utilize the saliency supervision, thereby enhancing the quality of generated VM- and TM-Summary. 

An appropriate evaluation metric is also crucial to the BiSSV task due to its multimodal output. 
Lin et al.~\cite{lin2023videoxum} propose using CLIPScore~\cite{hessel2021clipscore} to assess VM- and TM-Summary consistency. 
However, judging bimodal summaries by averaging image-text similarity ignores that the TM-Summary should align with visual elements according to their saliency.
Similarly, unimodal summarization metrics based on global ranking similarity~\cite{otani2019rethinking} treat all segments equally and overlook the need for summaries to prioritize the most salient parts. 
To address these issues, we introduce a novel metric, $NDCG_{MS}$, which considers saliency to jointly evaluate the outputs of different modalities in the BiSSV task.

In summary, our main contributions include:
\begin{itemize}
    \item We build a new large-scale dataset, BIDS, to support the investigation of bimodal summarization of videos.
    \item We propose an end-to-end framework UBiSS with a ranking-based optimization objective to generate bimodal summaries. Experiments demonstrate that our UBiSS outperforms multi-stage baselines.
    \item We propose a novel evaluation metric, $NDCG_{MS}$, to assess bimodal summaries more appropriately. 
\end{itemize}

\section{Related Work}
\label{sec: related work}

\begin{figure*}[t]
    \centering
    \includegraphics[width=0.9\linewidth]{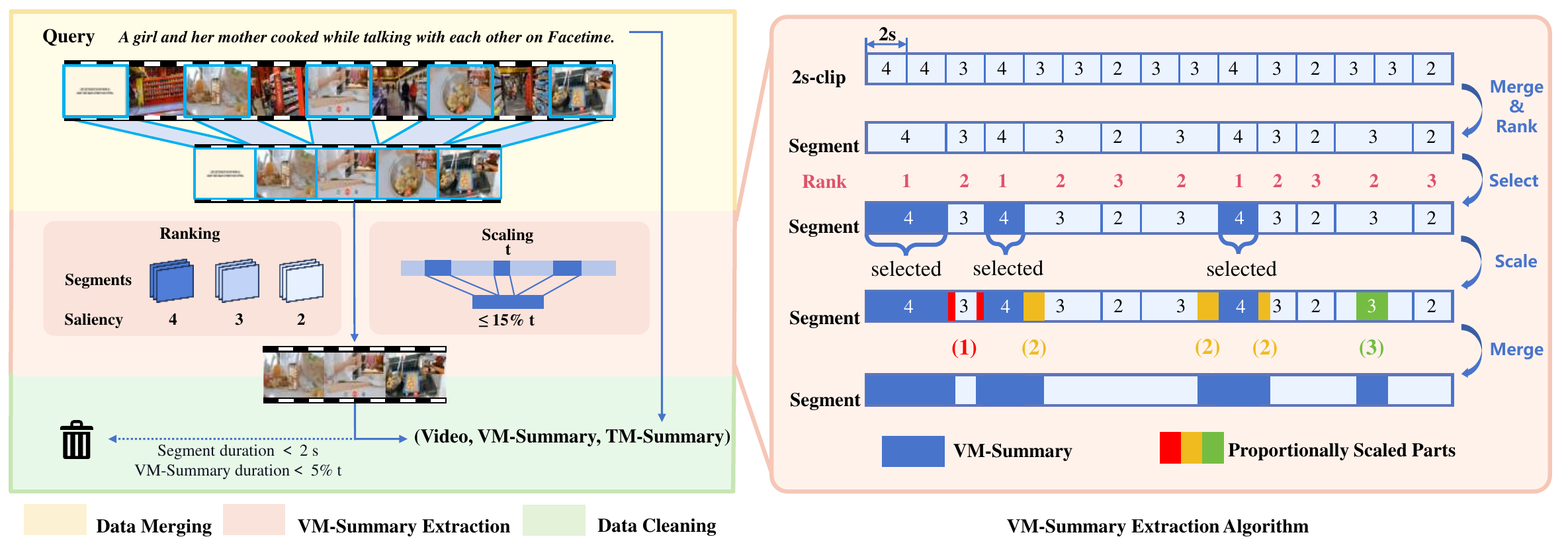}
    \vspace{-8pt}
    \caption{Illustration of the data processing procedure comprising data merging, VM-Summary extraction, and data cleaning. Each step is discussed in Section \ref{sec: datapro}. 
    The VM-Summary extraction algorithm is presented on the right side with colored numbers representing different scaling conditions: \textcolor{red}{(1) Both adjacent segments are selected.} \textcolor{myyellow}{(2) Only one adjacent segment is selected.} \textcolor{mygreen}{(3) No adjacent segments is selected.}
    }
    \label{fig: data_construct}
\end{figure*}

We can broadly divide related work on video summarization into two categories according to the output format: unimodal or multimodal video summarization. BiSSV belongs to the second category.

\textbf{Unimodal Video Summarization.} Traditional unimodal video summarization includes Textual-Modal Summarization of Videos (TMSV) and Visual-Modal Summarization of Videos(VMSV). Video captioning~\cite{venugopalan2015sequence} is a typical task in TMSV, which uses generic language to summarize videos, resulting in the loss of rich semantics. 
Dense video captioning~\cite{krishna2017dense}, on the other hand, aims to generate multiple captions with their corresponding time intervals for a given video. However, as dense captions focus on specific events within videos, they cannot provide an overview of the entire video. 
Traditional video summarization~\cite{gygli2014creating}, as a representative task in VMSV, aims to extract the most informative segments from videos. Due to the complex annotation process, mainstream video summarization datasets~\cite{gygli2014creating,song2015tvsum,de2011vsumm,chu2015video} are limited in size. 
Most video summarization datasets and approaches rely on the Knapsack algorithm to convert saliency scores into summarized segments~\cite{gygli2014creating,song2015tvsum,zhang2016video,lebron2018video,fajtl2019summarizing,8667390,ji2020deep,apostolidis2021combining}. 
However, Otani M et al.~\cite{otani2019rethinking} point out a bias in the Knapsack algorithm towards favoring shorter segments. 
Although VM-Summary offers more details than TM-Summary, it requires extended comprehension time.
In conclusion, unimodal video summarization inevitably results in semantic loss, while multimodal summarization leverages complementary modalities to provide concise yet informative summaries.

\textbf{Multimodal Video Summarization.} 
Chen et al.~\cite{chen2017video} first tries to improve the performance of both TMSV and VMSV. Nevertheless, the lack of paired summary annotations leads to low consistency between the two sub-tasks. 
Recently, Lin et al.~\cite{lin2023videoxum} introduce the VideoXum dataset and optimize TMSV and VMSV in a unified way.
However, the TM-Summary in VideoXum is generated by concatenating dense captions from ActivityNet Captions~\cite{krishna2017dense} with an average length of 49.9 words, which falls short of providing a succinct overview. 
The evaluation of multimodal video summarization also remains a challenge. 
Lin et al. ~\cite{lin2023videoxum} adopt average CLIPScore between VM- and TM-Summary~\cite{hessel2021clipscore} to assess consistency. 
However, it's essential for TM-Summary to align with visual elements according to their saliency.
To address these challenges, we construct a dataset, design a unified framework for bimodal summarization and propose $NDCG_{MS}$ to assess bimodal summaries jointly. 
\section{BIDS Dataset}
\label{sec: dataset}

As it is very costly to build a bimodal summarization dataset from scratch, we, therefore, leverage the QVHighlights dataset~\cite{lei2021detecting} to construct a \textbf{B}imodal V\textbf{ID}eo \textbf{S}ummarization dataset (\textbf{BIDS}) to support the investigation of the BiSSV task. The constructed BIDS dataset finally contains 8130 videos with corresponding ground-truth Visual-Modal (VM) and Textual-Modal (TM) Summaries and saliency scores annotated for each 2-second clip, indicating its significance. Following the restrictions of traditional video summarization~\cite{gygli2014creating}, we ensure that the length of the VM-Summary does not exceed 15\% of the original video's duration. We describe the data processing and analysis in detail in the following subsections.

\subsection{Data Processing}
\label{sec: datapro}

We aim to build a bimodal video summarization dataset with triplet data samples (video, TM-Summary, VM-summary), where the TM-Summary is a concise text description, and the VM-summary contains highlighted segments within the video. Firstly, we merge text-related segments from the original videos to guarantee that the TM-Summary accurately captures the main content of the video. 
Secondly, we design a ranking-based extraction algorithm to preserve the most salient visual content as VM-Summary. 
Lastly, we perform data cleaning to remove unsuitable videos that lack a clear focus for summarization. The overview of the BIDS building process is illustrated in Figure \ref{fig: data_construct}.

\noindent \textbf{Data merging.} QVHighlights~\cite{lei2021detecting} is a video dataset that supports query-based moment retrieval and highlight detection, with annotations of natural language query, segments relevant to the query, and saliency scores for each 2s-clip within the segments. Taking the query as the TM-Summary, we merge the relevant segments chronologically as original videos in our dataset. In this way, we obtain a (video, TM-Summary) pair, for which we subsequently extract the VM-Summary. 

\noindent \textbf{VM-Summary extraction.} We utilize the annotated 2s-saliency scores for VM-Summary extraction. 
Unlike the Knapsack algorithm utilized by previous video summarization datasets~\cite{song2015tvsum,gygli2014creating}, our extraction algorithm retains salient visual content within long segments and avoids favoring short segments. An illustration of this algorithm is presented in Figure ~\ref{fig: data_construct}. 
We also provide a  pseudo-code in Appendix \ref{sec: pseudo code}.

(a) \textit{Ranking}. We first merge adjacent 2s-clips with the same saliency scores into segments. Then, we rank all the candidate segments according to their saliency scores. The candidate segments are subsequently selected for VM-Summary in descending order. To comply with the length limit of VM-Summary (15\% video duration in our case), we may need to scale some candidate segments.

(b) \textit{Scaling}. As the candidate segments vary in length, the purpose of scaling is to preserve informative parts within segments while guaranteeing conciseness. Specifically, candidate segments with the same score will be appended to the VM-Summary if it does not surpass the length constraint. Otherwise, these segments are proportionally scaled. 
We assume that the parts closer to higher-scored segments usually contain more valuable information.
Therefore, if the segment has higher-scored neighbors, adjacent parts closer to those neighbors are preserved (colored in \textcolor{red}{red} and \textcolor{myyellow}{yellow}, indicating two and one higher-scored neighbors, respectively); otherwise, its central part is preserved (colored in \textcolor{mygreen}{green}). The scaled segments are appended to the VM-Summary, and the segments with lower ranks are all rejected.

\noindent \textbf{Data cleaning.} Finally, we remove segments shorter than 2 seconds and videos with VM-Summary occupying less than 5\% of the video duration since they lack clear focal points for summarization. Finally, of 8,172 videos, only 42 (0.51\%) videos are removed.

\subsection{Data Analysis}
\label{sec: dataana}

Traditional video summarization datasets use the Knapsack algorithm to generate VM-Summary~\cite{gygli2014creating,song2015tvsum}. 
However, Otami M et al.~\cite{otani2019rethinking} point out that their segmentation-selection pipeline favors short segments since selecting long segments costs more. 
However, long and visually consistent segments can also contain informative moments. For example, when watching a video of \textit{someone playing basketball}, most of the visual content is similar, but we can still identify key moments, such as \textit{shooting}.
Inspired by humans' ability to distinguish important moments in long videos, we choose to scale the candidate segments instead of rejecting them entirely. As a result, our VM-Summary shows a stronger correlation between the saliency scores and the selected segments.  

We use Spearman's correlation coefficient $\rho$~\cite{zwillinger1999crc} to validate the effectiveness of our VM-Summary extraction algorithm. A higher coefficient between the saliency scores $S$ and the frame-level selection sequence $F$ (1 for the frame being selected into the VM-Summary and 0 for otherwise) indicates more salient content is preserved, which is the goal of summarization. 
As presented in Table ~\ref{tab: dataset comparison}, BIDS has the highest Spearman's $\rho$ compared to traditional datasets. Moreover, Spearman's $\rho$ between $S$ and $F$ (generated by annotators) surpasses the $\rho$ between $S$ and GT-$F$ (obtained by applying Knapsack algorithm over the annotated saliency scores) in SumMe~\cite{gygli2014creating}, which further demonstrates that Knapsack algorithm can not effectively preserve salient parts within long segments. 

\begin{figure}[t]
    \centering
    \includegraphics[width=0.95\linewidth]{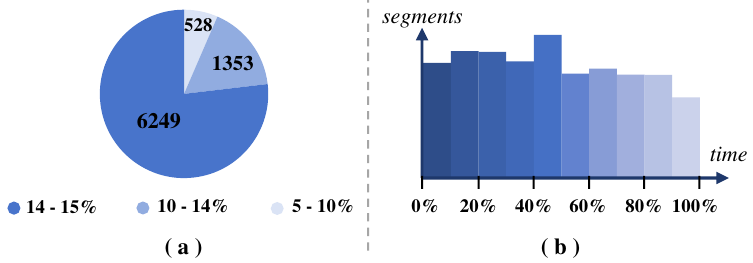}
    \vspace{-6pt}
    \caption{(a) Distribution of duration ratio between VM-Summary and original video; (b) Distribution of temporal positions of the segments selected into the VM-Summary in the original video.
    }
    \label{fig: distribution}
\end{figure}

After removing invalid and duplicate videos, BIDS contains 8130 videos, with 5854/650/1626 videos for training/validation/test set. 
We ensure that the original videos between different sets do not overlap to avoid data leakage.  
The data statistics of BIDS are presented in Table \ref{tab: dataset statistics}. As presented in Figure \ref{fig: distribution}, our algorithm is able to generate VM-summaries within a strict length constraint, with the majority occupying 14-15\% of the video's duration. Furthermore, the segments in a VM-Summary are evenly distributed throughout the corresponding video.

\begin{table}[t]
    \centering
    \small  
    \caption{Comparison with traditional video summarization datasets.
    $\rho$: Average Spearman's correlation coefficient. 
    Sig.: Significance (p < 0.05). 
    $S$: Saliency score.
    $F$: Frame-level sequence indicating each frame is selected (1) or not selected (0) into the VM-Summary. 
    GT-$F$: the $F$ is calculated by averaging human annotated scores for each video in SumMe~\cite{gygli2014creating} and TVSum~\cite{song2015tvsum}.
    dp: the $F$ is obtained by the Knapsack algorithm.
    } 
    
    \vspace{-8pt}
    \begin{tabular}{ccccc}
    \toprule
    \textbf{Dataset}                & \textbf{Set of Variables}          & $\boldsymbol{\rho}$ & \textbf{Sig.}  & \textbf{\# of Videos}    \\
    \midrule
    \multirow{2}{*}{SumMe~\cite{gygli2014creating}} &($S$, GT-$F_{dp}$) & 0.34                               & \checkmark & \multirow{2}{*}{25} \\
                           &($S$, $F$)        &  \underline{0.44}                        & \checkmark &                     \\
                           \midrule
    \multirow{2}{*}{TVSum~\cite{song2015tvsum}} &($S$, GT-$F_{dp}$)& 0.31                               & \checkmark & \multirow{2}{*}{50} \\
                           &($S$, $F_{dp}$)    & 0.24                               & $\times$   &                     \\
                           \midrule
    BIDS(ours)           &($S$, GT-$F$)     & \textbf{0.52}                      & \checkmark & \textbf{8130}     \\ 
    \bottomrule
    \end{tabular}
    \label{tab: dataset comparison}
\end{table}

\begin{figure*}[t]
    \centering
    \includegraphics[width=0.8\linewidth]{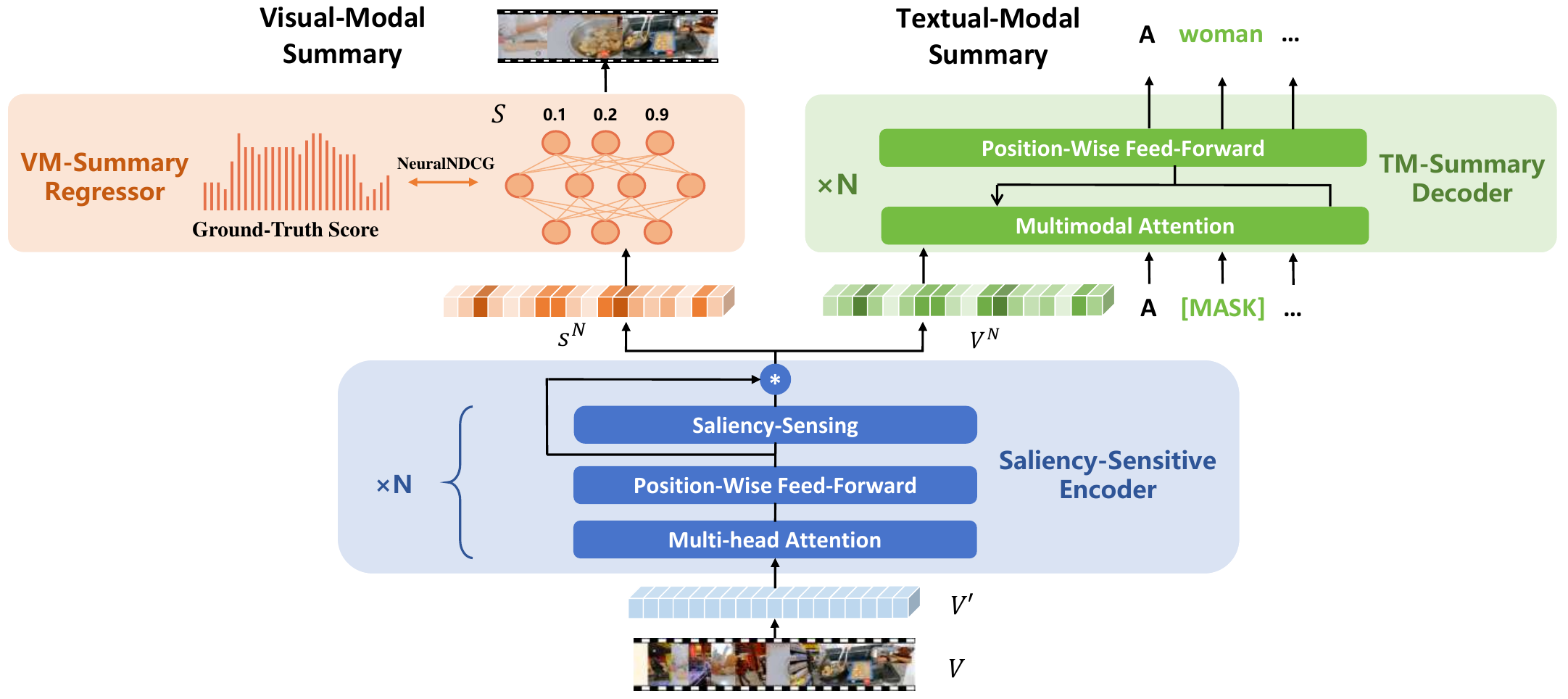}
    \vspace{-8pt}
    \caption{Model Architecture of UBiSS. 
    }
    \label{fig: framework}
\end{figure*}

\begin{table*}[t]
    \centering
     \small
    \captionsetup{skip=10pt}
    \caption{Statistics of BIDS. 
    {VM: Visual-Modal Summary. TM: Textual-Modal Summary.}
    \vspace{-8pt}
    }
    \begin{tabular}{ccccccc}
           \toprule
           & \textbf{Avg. Video Len(s)} & \textbf{Total Video Len(h)} & \textbf{Avg. VM Len(s)} & \textbf{Avg. VM proportion(\%)} & \textbf{Avg. TM Len(word)} & \textbf{\# of Videos} \\
           \midrule
            Training   & 43.55           & 70.82             & 6.05         & 14.07               & 10.52            & 5854             \\
            Validation & 40.05           & 7.23              & 5.57         & 14.07               & 10.41            & 650              \\
            Test       & 44.83           & 20.25             & 6.19         & 14.12               & 10.42            & 1626             \\
            All        & 43.53           & 98.3              & 6.04         & 14.08               & 10.49            & 8130      \\
            \bottomrule
    \end{tabular}
    \label{tab: dataset statistics}
\end{table*}

\section{Method}
\label{sec: method}

For a given video, the BiSSV task aims to generate a textual-modal summary (TM-Summary) and extract the most related clips to form a visual-modal summary (VM-Summary). We first introduce the necessary symbols and then present our proposed \textbf{U}nified framework for \textbf{Bi}modal \textbf{S}emantic \textbf{S}ummarization of videos, \textbf{\textit{UBiSS}}, along with our training strategy and the proposed evaluation metric in the subsequent sections.

Let us denote the input video as $V=\left\{v_i\right\}, i \in[0, T], v_i \in R^{C \times h \times w}$, where $T$ represents the number of frames, and $C, h, w$ refer to the channel, height, and width of video frames, respectively. The goal of BiSSV is to generate the TM-Summary $W=\left\{w_i\right\}, i \in[0, l]$ that globally summarizes the video, and a VM-Summary represented by a frame-level selection sequence $F=\left\{f_i\right\}, i \in [0, T], f_i \in\{0,1\}$, indicating whether each frame is selected for the summary.

One might intuitively think of the BiSSV task as a combination of two subtasks: visual- and textual-modal summarization. A straightforward solution is to combine techniques for these two subtasks. However, such a pipeline-type solution is not optimal because the output VM- and TM-Summaries in BiSSV are closely related, so they should be tightly coupled during the generation process.

Inspired by unified end-to-end frameworks~\cite{liu2022umt,jiang2022joint}, we propose \textbf{\textit{UBiSS}}, a unified solution that models bimodal summaries simultaneously. By reformatting VM-Summary extraction as a sequence recommendation problem, we further adopt a ranking-based optimization objective~\cite{Pobrotyn2021NeuralNDCG} for saliency learning. 

\subsection{Model Architecture}

Figure \ref{fig: framework} depicts the overall model architecture of UBiSS, which consists of three main modules: saliency-sensitive encoder, TM-Summary decoder, and VM-Summary regressor.  

\textbf{Saliency-sensitive encoder.} 
The input video is first embedded as a feature sequence $V^{\prime}=\left\{v_i^{\prime}\right\}, i \in [0, t], v^{\prime}_i \in R^d$ via a pretrained model~\cite{szegedy2015going,liu2022video}, where $t$ is the length of the sequence, and $d$ represents the feature dimension. $V^{\prime}$ is then fed into the saliency-sensitive encoder. Following \cite{song2021towards}, each encoder layer contains a saliency-sensing layer for learning temporal saliency information, except for the traditional multi-head attention layer followed by a position-wise feed-forward layer. The saliency-sensing layer first calculates the sigmoid function $s^{i}$ based on the output of the feed-forward layer. Then, it uses $s^{i}$ as weights to scale the input. In this way, the importance of visual content is considered during temporal modelling. Its calculation process is as follows, and all layers included do not change the dimension of the input vector:
\vspace{-6pt}
\begin{align}
    \centering
     \overline{V}^i&=V^{i-1}+MultiHeadAttention(V^{i-1}) \\
     s^i&=\sigma(PositionwiseFeedForward(\overline{V^i})) \\
      V^{i}&=s^{i}{\cdot}\overline{V^{i}} 
\end{align}
where $\sigma$ denotes the sigmoid function, and $s^{i}$ represents the saliency of the visual features output by the $i$-th layer of the encoder.

\textbf{VM-Summary regressor.} Two linear layers transform the score $s^N$ from the last saliency-sensing layer into the predicted saliency score $S$. Scores are then transformed into $F$ using the same extraction algorithm described in Section \ref{sec: datapro}.

\textbf{TM-Summary decoder.} The TM-Summary decoder~\cite{devlin2018bert} utilizes the input visual features $V^N$ and textual tokens to generate TM-Summary. Its attention layer is constrained so that textual tokens can only attend to previously generated tokens, and visual features cannot attend to textual features. During training, masked ground-truth TM-Summary is used as input. During inference, the decoder generates TM-Summary in an auto-regressive manner.

\subsection{Training Strategy}
\label{sec: training}

Following previous approaches~\cite{lin2022swinbert}, we employ masked language modelling ($L_{MLM}$) as the optimization objective for textual-modal summarization. A specific percentage of the ground-truth TM-Summary is replaced with a [MASK] token, and the model is required to predict these masked tokens. This objective effectively drives the model to learn cross-modality representations.

Inspired by pair-wise ranking networks for personalized video summarization~\cite{yao2016highlight,saquil2021multiple}, we further optimize visual-modal summarization by reformatting it as a learning-to-rank problem.
Instead of predicting the precise saliency score, we supervise the model to learn the appropriate score ranking of video frames. 
To achieve this goal, we employ a list-wise ranking-based optimization objective neuralNDCG($L_{RBL}$)~\cite{Pobrotyn2021NeuralNDCG} to learn global saliency rankings. This objective aims to increase the value of a list-wise similarity metric Normalized Discounted Cumulative Gain (NDCG)~\cite{jarvelin2002cumulated}, which is calculated as follows:
\vspace{-0pt}
\begin{align}
    \centering
    DCG@k=&\sum_{j=1}^k\frac{2^{\overline{s_j}}-1}{\log_2(j+1)} \label{eq: ndcg}\\
    NDCG@k=&\frac1{maxDCG@k}DCG@k
\end{align}
where $j$ represents the ranking of a visual feature in the model's prediction; $\overline{s_j}$ represents the ground-truth saliency score of the $j$-th ranked feature; $maxDCG@k$ denotes the $DCG@k$ when the model's output ranking is identical to the ground-truth; $k$ indicates the number of elements taken into account (the top k visual features, in our case).

Our overall optimization objective is the combination of $L_{MLM}$ and $L_{RBL}$. We observe that learning textual-modal summarization takes longer than learning temporal saliency.
Therefore, we first only optimize $L_{MLM}$ for N epochs as a warm-up for textual-modal summarization, then simultaneously train UBiSS with $L_{MLM}+L_{RBL}$. 
More details about the texual-modal summarization warm-up are described in Appendix \ref{sec: warm up}.

\subsection{Evaluation of BiSSV}

As the BiSSV task involves multimodal output, a global evaluation metric is needed to evaluate the overall performance of different approaches. 

\begin{figure}[t]
    \centering
    \includegraphics[width=0.95\linewidth]{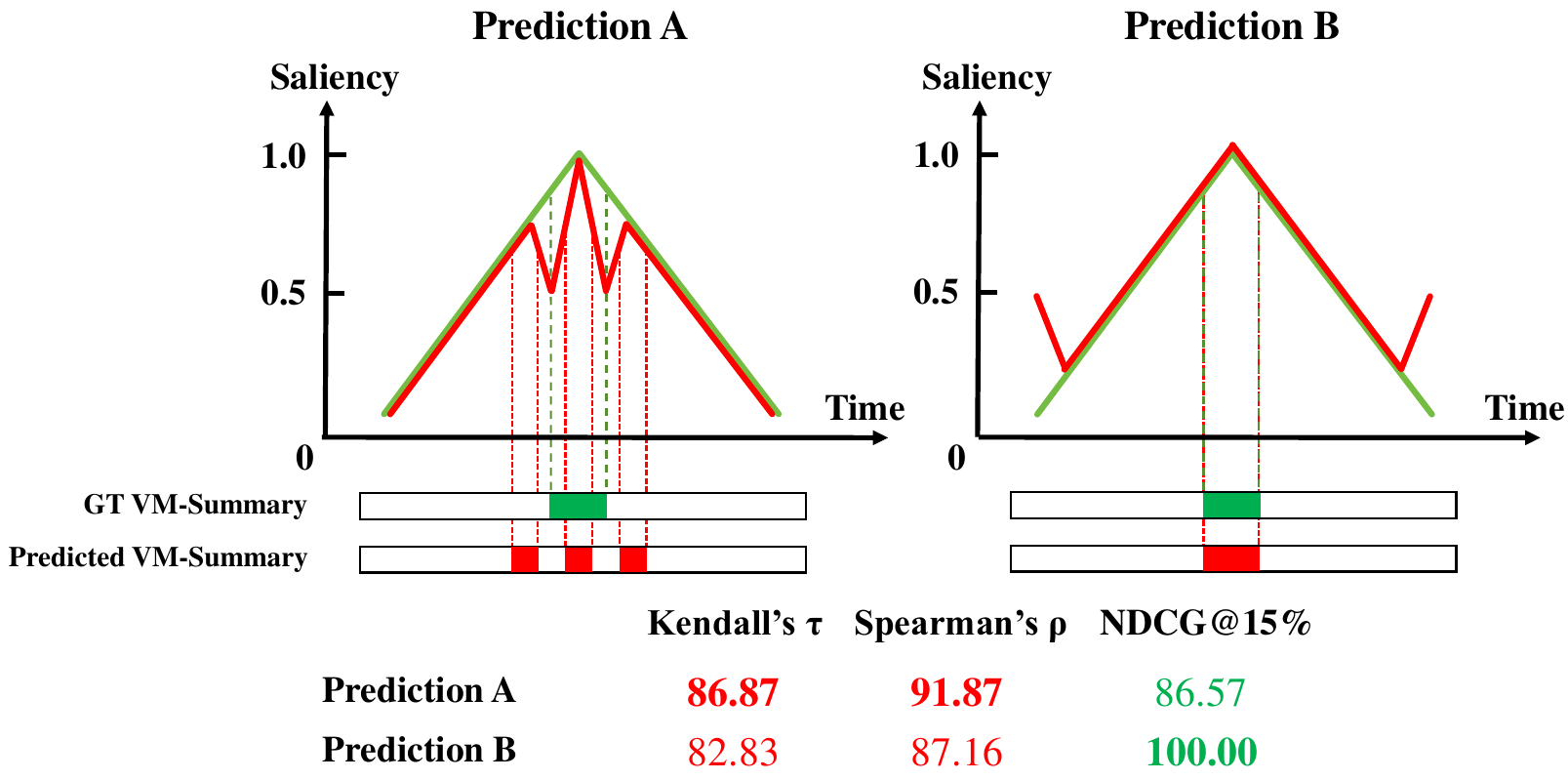}
    \vspace{-10pt}
    \caption{Comparison of metrics for ranking similarity evlation~\cite{kendall1945treatment,zwillinger1999crc,jarvelin2002cumulated}. 
    {Prediction A} makes an incorrect prediction on two highest-scored segments, while {Prediction B} makes an incorrect prediction on two lowest-scored segments.
    Since Kendall's $\tau$ and Spearman's $\rho$ treat all segments equally while assessing ranking similarity, they both favor Prediction A. However, the incorrect prediction of A results in an inaccurate VM-Summary.
    Our proposed metric, NDCG@15\%, prioritizes the ranking similarity of most salient segments and favors Prediction B more.
    }
    \label{fig: ndcg}
\end{figure}

We first explore visual-modal summarization metrics for inspiration. 
One way to evaluate VM-summary is to assess global ranking similarity, as proposed by ~\cite{otani2019rethinking}, which measures the alignment of the ground-truth and the predicted saliency score sequences. 
However, these metrics overlook the inherent inequality among segments during the summarization process. Given the direct correlation between score prediction and VM-Summary extraction, more salient segments should have a more significant influence on the evaluation result. 
However, previous metrics~\cite{kendall1945treatment,zwillinger1999crc} that measure global ranking similarity treat segments equally. A quantitative example is illustrated in Figure \ref{fig: ndcg}, where Prediction A incorrectly predicts the highest-scored segments. In contrast, Prediction B makes an incorrect prediction on the lowest-scored segments. Previous metrics, including Kendall's $\tau$~\cite{kendall1945treatment} and Spearman's $\rho$~\cite{zwillinger1999crc}, favor Prediction A more, though its mistaken prediction of most salient segments directly leads to inaccurate VM-Summary. 
Only our proposed NDCG@15\%, taking ground-truth saliency as weights for different segments, can distinguish Prediction B's superior performance in prioritizing higher-scored segments. We introduce NDCG in detail as follows.

\begin{table*}[htbp]
    \centering
    \small
    \captionsetup{skip=10pt}
    \caption{Comparison to multi-stage unimodal baselines. The best and second-best results are \textbf{bolded} and \underline{underlined}. {CNN/Swin: CNN~\cite{szegedy2015going} or Video Swin Transformer~\cite{liu2022video} feature. N/M: applying the NeuralNDCG~\cite{Pobrotyn2021NeuralNDCG} or MSE loss. VT-C: VT-CLIPScore~\cite{lin2023videoxum}}.
    }
    \vspace{-8pt}
    \begin{tabular}{c|lccccccccc}
    \toprule
             & \multirow{2}{*}{\textbf{Method}}      & \multirow{2}{*}{\textbf{CIDEr}} & \multirow{2}{*}{$\tau$} & \multirow{2}{*}{\textbf{VT-C}} & \multicolumn{2}{c}{\textbf{\textit{Visual Modality}}}       & \multicolumn{2}{c}{\textit{\textbf{Textual Modality}}}        & \multicolumn{2}{c}{\textit{\textbf{Bimodal Summarization}}}         \\
                    \cmidrule(lr){6-7}  \cmidrule(lr){8-9} \cmidrule(lr){10-11}
           &        &                     &   &                      & \textbf{NDCG@15\%}      & \textbf{NDCG@all}       & \textbf{NDCG@15\%}      & \textbf{NDCG@all}       & \textbf{NDCG@15\%}      & \textbf{NDCG@all}       \\ \hline
  1&  PGL-Swin       & \underline{40.71}                  & 15.23      & \textbf{24.26}          & 64.74          & 84.32          & 56.91          & \underline{81.37}          & 60.83          & 82.85          \\
   2& Swin-PGL       & \textbf{45.24}         & 14.80      & \underline{24.19}             & 65.01          & 84.43          & \textbf{57.70}    & \textbf{81.67} & 61.35          & 83.05          \\
   3& UBiSS(CNN, N)  & 30.43                  & 14.54      & 22.70                      & 65.32          & 84.47          & 55.84          & 80.90          & 60.58          & 82.69          \\
   4& UBiSS(Swin, M) & 39.81                  & \underline{17.58}  & 23.84        & \underline{66.30}    & \underline{84.89}    & \underline{57.11}        & 81.35          & \textbf{61.71}    & \underline{83.12}    \\
   5& UBiSS(Swin, N) & 39.87     & \textbf{17.69}  & 23.68     & \textbf{66.71} & \textbf{85.06} & 56.61 & 81.26    & \underline{61.66} & \textbf{83.16} \\ 
    \bottomrule
    \end{tabular}
    \label{tab: pipeline}
\end{table*}

By assigning weights to ground-truth scores based on predicted saliency ranking (Eq. (\ref{eq: ndcg})), NDCG naturally gives greater significance to segments with higher ground-truth saliency.
Therefore, we employ NDCG~\cite{jarvelin2002cumulated}  for bimodal summarization evaluation. 
We use NDCG@15\% and NDCG@all to represent the ranking similarity of the top 15\% and all frames, respectively. In this way, NDCG@15\% directly assesses the model's ability to prioritize the most salient segments, while NDCG@all evaluates saliency-weighted global ranking similarity.

The NDCG between the model's predicted scores and ground-truth saliency is denoted as $NDCG_{VM}$ for visual-modal summarization evaluation. 
We modify NDCG to evaluate the textual-modal summarization by applying vision-language pretraining models. Specifically, we first calculate the similarity between TM-Summary and each frame. Since the TM-Summary should align with different parts of the video according to their saliency, we calculate $NDCG_{TM}$ between this similarity sequence and the ground-truth saliency sequence. The average score of $NDCG_{VM}$ and $NDCG_{TM}$, annotated as $NDCG_{MS}$, represents bimodal summarization performance. We calculate NDCG after score normalization. We also report VT-CLIPScore~\cite{hessel2021clipscore} for cross-modal consistency~\cite{lin2023videoxum}. 

Additionally, we conduct a breakdown assessment of the VM- and TM-Summary using the corresponding standard metrics, respectively. For textual-modal summarization, BLEU-4(B4)~\cite{papineni2002bleu}, SPICE (S)~\cite{anderson2016spice}, ROUGE-L(R-L)~\cite{lin2004rouge}, CIDEr(C)~\cite{vedantam2015cider}, and METEOR(M)~\cite{banerjee2005meteor} are reported. For visual-modal summarization, F-score between ground-truth and predicted $F$, Spearman's $\rho$~\cite{zwillinger1999crc} and Kendall's $\tau$~\cite{kendall1945treatment} are reported. 
\section{Experiments}
\label{sec: experiments}
\begin{figure*}[ht]
    \centering
    \includegraphics[width=0.9\linewidth]{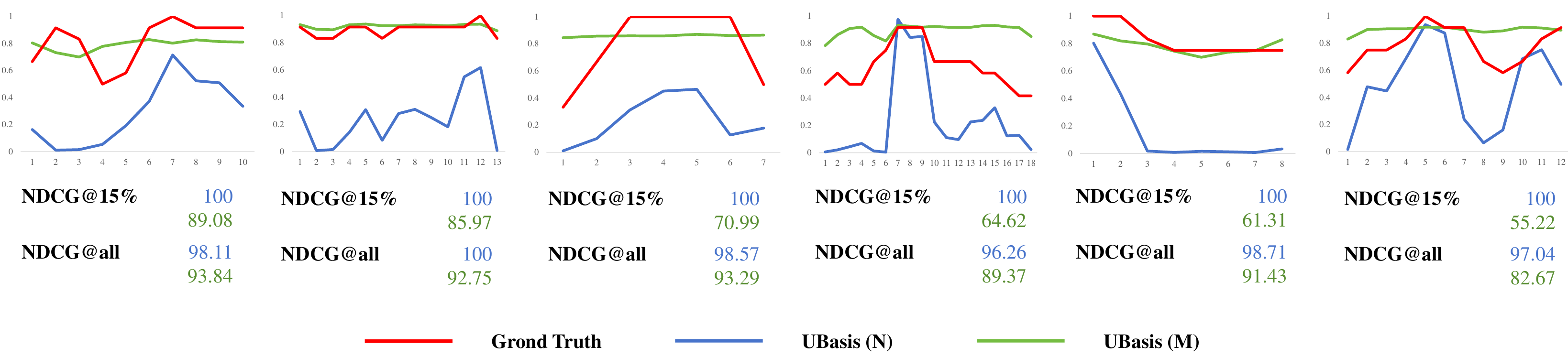}
    \vspace{-8pt}
    \caption{Visualization of the results of models trained with different loss functions. 
    $NDCG_{VM}$ for each video is presented.
    \emph{N: NeuralNDCG~\cite{Pobrotyn2021NeuralNDCG}. M: Mean Square Error.}
    }
    \label{fig: loss}
\end{figure*}

\subsection{Implemention Details}

Videos are downsampled to 8 fps. We use the ImageNet-1k pre-trained GoogleNet~\cite{szegedy2015going} pool (5b) layer and Video Swin Transformer~\cite{liu2022video} for feature extraction. For GoogleNet, the mean pooling of features every 2 seconds is used as the pre-extracted video features. The feature dimension is 1024. The saliency-sensitive encoder has an embedding dimension of 1024, an intermediate layer dimension of 2048, and 4 encoder layers with 4 attention heads. The TM-Summary decoder has a maximum text sequence length of 45 and is implemented based on the Hugging Face Transformer\footnote{https://github.com/huggingface/transformers}. The VM-Summary regressor employs a dropout probability of $p$=$0.5$. We use the AdamW optimizer~\cite{loshchilov2018decoupled} and the MultiStepLR learning rate scheduler. Textual-modal summarization warm-up epoch N is set to 30. We use CLIP-ViT-B/32~\cite{radford2021learning} as the vision-language pre-training model to calculate $NDCG_{MS}$. Unless otherwise specified, the test results are reported after training for 200 epochs with batch size 32, and the model is selected based on CIDEr~\cite{vedantam2015cider} of the validation set. We report the average results of five runs with random seeds.

\subsection{Comparison to Multi-stage Baselines} 
\label{sec: multi-stage}

Table \ref{tab: pipeline} presents the comparison of our UBiSS with multi-stage baselines (Swin-PGL and PGL-Swin) that simply integrate two state-of-the-art unimodal summarization methods: PGL-SUM~\cite{apostolidis2021combining} is a traditional visual-modal summarization model that utilizes global and local attention with positional encoding to identify highlighted moments in videos. SwinBERT~\cite{lin2022swinbert} is the first end-to-end video captioning model that updates both the visual extractor and text generator during training. 
For Swin-PGL, we train a linear layer that projects the output token embeddings of SwinBERT~\cite{lin2022swinbert} into the same dimension as the visual features and concatenates them to train PGL-SUM~\cite{apostolidis2021combining}. For PGL-Swin, we use VM-Summaries extracted by PGL-SUM~\cite{apostolidis2021combining} to train and test SwinBERT~\cite{lin2022swinbert}. The pretrained SwinBERT and PGL-SUM are selected based on CIDEr~\cite{vedantam2015cider} and Kendall's $\tau$~\cite{kendall1945treatment}, respectively. 

As shown in Table \ref{tab: pipeline}, using only VM-Summary as video inputs results in diminished performance in the textual-modal summarization for PGL-Swin. Swin-PGL, utilizing a robust textual-modal summarization baseline~\cite{lin2022swinbert} that incorporates an online visual extractor, attains the highest $NDCG_{TM}$ score. However, the performance of Swin-PGL lags behind UBiSS in visual-modal summarization, suggesting that a straightforward fusion method fails to leverage the connections between modalities thoroughly.

On cross-modal consistency, UBiSS achieves comparable performance to Swin-PGL in VT-CLIPScore~\cite{lin2023videoxum}. However, it is worth noting that simply averaging image-text similarity can not adequately represent summarization capability. As evident from higher $NDCG_{MS}$, UBiSS achieves better performance in bimodal summarization than pipeline-type solutions.

From Table \ref{tab: pipeline}, UBiSS with a ranking-based optimization objective (row 5) surpasses the traditional regression loss model (row 4) in visual-modal summarization, indicating the effectiveness of our optimization objective in enhancing saliency modeling. However, in textual-modal summarization, UBiSS trained with the traditional regression loss outperforms that with the ranking-based objective. This decline in textual-modal summarization performance can primarily be attributed to the calculation process of NDCG, during which less salient visual content is assigned lower weights. Consequently, the model may resort to taking shortcuts by assigning extremely low scores to less critical visual features during optimization, as depicted in Figure ~\ref{fig: loss}. Such biased visual features can negatively impact the textual-modal summarization capabilities.
\begin{table}[t]
    \centering
     \small
    \caption{Comparison to visual-modal summarization models. All models are trained from scratch. All models, except ours, use mean square error as loss function. {CNN/Swin refer to the CNN~\cite{szegedy2015going} or Video Swin Transformer~\cite{liu2022video} feature.} The best and second-best results are \textbf{bolded} and \underline{underlined}.}
    \vspace{-8pt}
    \begin{tabular}{cccccccc}
        \toprule
         & \textbf{Feature} & \textbf{F-score}           & \textbf{$\boldsymbol{\tau}$}         & \textbf{$\boldsymbol{\rho}$} \\
         \midrule
    Random           & CNN     & 13.85       & 0.37           & 0.49      \\
    Linear           & CNN     & 17.78       & 13.27          & 16.74             \\
    Transformer      & CNN     & 17.29       & 10.74          & 13.57                \\
    PGL-SUM~\cite{apostolidis2021combining}          & CNN     & 19.03       & 15.23          & 19.16             \\
    UBiSS & CNN     & \underline{19.67}    & \underline{16.17}   & \underline{20.41}    \\
    UBiSS & Swin    & \textbf{19.96} & \textbf{18.04}      & \textbf{22.55}\\
        \bottomrule
    
    \end{tabular}
    \label{tab: video summarization}
\end{table}

\begin{table}[t]
    \small
    \caption{Comparison to textual-modal summarization (a.k.a. video captioning) models. {CNN/Swin refer to the CNN~\cite{szegedy2015going} or Video Swin Transformer~\cite{liu2022video} feature.} The best and second-best results are \textbf{bolded} and \underline{underlined}. }
    \vspace{-8pt}
    \centering
    \begin{tabular}{cccccccc}
    \toprule
                           & \textbf{Params.} & \textbf{B4}        & \textbf{M}         & \textbf{R-L}        & \textbf{C}          & \textbf{S}          \\
    \midrule
    SwinBERT~\cite{lin2022swinbert}             & 224 M      & \textbf{5.07} & \textbf{10.88} & \textbf{24.77} & \textbf{46.81} & \textbf{15.60} \\
    UBiSS(CNN)   & 159 M      & 3.42          & 8.53           & 19.68         & 30.43          & 10.23          \\
    UBiSS(Swin)  & 159 M      & \underline{4.29}    & \underline{9.96}     & \underline{22.28}    & \underline{39.87}    & \underline{13.02}   \\
    \bottomrule
    \end{tabular}
    
    \label{tab: video captioning}
\end{table}

\subsection{Comparison to Unimodal Summarization}
\begin{figure*}[t]
    \centering
    \includegraphics[width=\linewidth]{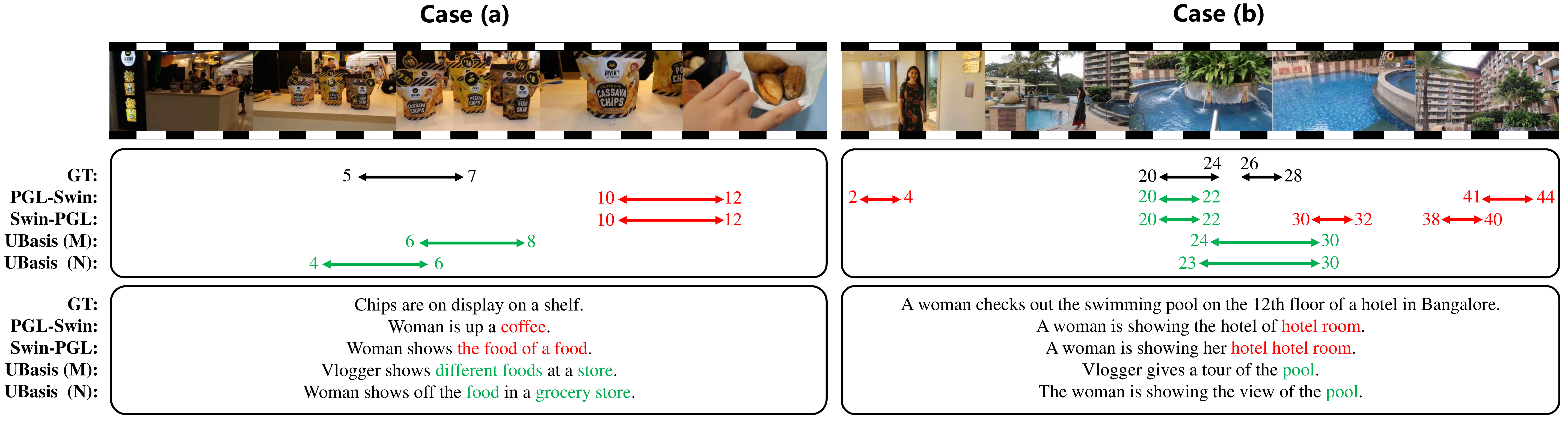}
    \vspace{-10pt}
    \caption{Visualization of bimodal summaries generated by pipeline-type baselines and UBiSS. \textcolor{red}{Red} represents irrelevant or incorrect content in the TM-Summary or a lack of overlap between the current segment and the ground truth segments in the VM-Summary. \textcolor{mygreen}{Green} represents accurate content in the TM-Summary or overlap between the current segment and the ground truth segments in the VM-Summary.}
    \label{fig: case}
\end{figure*}

\begin{table*}[t]
    \centering
    \caption{Loss ablation. $N_{VM}$ and $N_{TM}$ refers to $NDCG_{VM}$ and $NDCG_{TM}$. The best results are \textbf{bolded}.}
    \vspace{-8pt}
    \small
    \begin{tabular}{ccccccccccccc}
    \toprule
    \multicolumn{1}{l}{} & \multicolumn{5}{c}{\textbf{\textit{Visual-Modal Summarization}}}                                                           & \multicolumn{7}{c}{\textbf{\textit{Textual-Modal Summarization}}}                                   \\ \cmidrule(r){2-6}  \cmidrule(r){7-13}
                         & \textbf{F-score}           & $\boldsymbol{\tau}$         & $\boldsymbol{\rho}$         & $\mathbf{N_{VM}@15\%}$      & $\mathbf{N_{VM}@all}$     & \textbf{B4}        & \textbf{M}         & \textbf{R-L}        & \textbf{C}          & \textbf{S}      &$\mathbf{N_{TM}@15\%}$ &$\mathbf{N_{TM}@all}$  \\ 
    \midrule
    MSE                  & 19.48                   & 17.52          & 21.95          & 66.22          & 84.87        & 4.06 & 9.87           & 22.21       & 39.81 & \textbf{13.61}  &\textbf{56.61}  &\textbf{81.26} \\
    NeuralNDCG           & \textbf{19.96}  & \textbf{18.04} & \textbf{22.55} & \textbf{66.77} & \textbf{85.07}  & \textbf{4.29}         & \textbf{9.96} & \textbf{22.28} & \textbf{39.87}          & {13.02}     &{55.84}   &{80.90}  \\
    \bottomrule
    \end{tabular}
    \label{tab: loss ablation}
\end{table*}

We also compare UBiSS with unimodal summarization models. The comparison to visual-modal summarization model PGL-SUM~\cite{apostolidis2021combining} and other basic models are presented in Table \ref{tab: video summarization}, all chosen by Kendall's $\tau$~\cite{kendall1945treatment}. All models reported are trained with 2s-feature. 
Since F-score is significantly influenced by the post-processing algorithm\cite{otani2019rethinking}, we obtain the final VM-Summary by the VM-Summary extraction algorithm introduced in Section \ref{sec: datapro} to ensure fair comparison. From Table \ref{tab: video summarization}, UBiSS outperforms other baselines in terms of both interval overlap (F-score) and global ranking similarity (Kendall's $\tau$~\cite{kendall1945treatment} and Spearman's $\rho$~\cite{zwillinger1999crc}), as our model could learn saliency information from the integration of different modalities. 

The comparison results between UBiSS and the end-to-end video captioning model SwinBERT~\cite{lin2022swinbert} are presented in Table \ref{tab: video captioning}. The comparison is not entirely fair since UBiSS utilizes pre-extracted visual features and has significantly fewer learnable parameters than SwinBERT, which relies on an online feature extractor and requires substantial computational resources. For instance, the 32-frame version of SwinBERT consumes approximately 16 times more GPU memory during training compared to UBiSS with the same batch size. We observed significant improvements with better visual features for UBiSS, indicating a potential direction for advancement. 

\begin{table}[t]
\caption{Human evaluation of different forms of summaries. 
Satis/Inform refer to satisfaction and informativeness.}
\vspace{-8pt}
\centering
\begin{tabular}{lcc}
\toprule
                & \textbf{Satis} & \textbf{Inform} \\
\midrule
TM-Summary Only      & 3.55                      & 3.42                       \\
VM-Summary Only     & \underline{3.78}                      & \underline{3.75}                       \\
Bimodal Summary & \textbf{4.14}                      & \textbf{4.25}               \\       
\bottomrule
\end{tabular}
\label{tab: summary form}
\end{table}

\subsection{Loss Comparison}

The comparison between UBiSS trained with different losses is shown in Table \ref{tab: loss ablation}. In contrast to the traditional Mean Square Error (MSE) regression objective, we empirically observe that employing a list-wise ranking-based objective~\cite{Pobrotyn2021NeuralNDCG} enables the model to grasp the relative relationships among video frames. Consequently, UBiSS (NeuralNDCG) demonstrates improvements across the majority of metrics, indicating that our optimization objective can effectively enhance performance in both traditional and our proposed metrics.

Specifically, models trained with MSE may take shortcuts by pushing the average score of the model's prediction toward the average score of the ground-truth data. Figure \ref{fig: loss} also shows that the model trained with MSE tends to produce overly smoothed prediction, resulting in a lower $NDCG_{VM}$ score. In contrast, the model trained with a ranking-based loss is much more sensitive to temporal saliency changes. 

\begin{table}[t]
\caption{Human evaluation of bimodal summaries of UBiSS and pipeline-type baselines. Acc refers to accuracy.}
\vspace{-8pt}
\begin{tabular}{lccc}
\toprule
          & \textbf{Acc(TM)}  & \textbf{Acc(VM)}  & \textbf{Consistency}     \\
\midrule
PGL-Swin  & 2.98          & 3.96          & 3.07          \\
Swin-PGL  & 2.85          & 3.94          & 2.90           \\
UBiSS(M)(ours) & \underline{3.48}          & \underline{4.06}          & \underline{3.52}          \\
UBiSS(N)(ours) & \textbf{3.99} & \textbf{4.20} & \textbf{4.06}\\
\bottomrule
\end{tabular}
\label{tab: baseline human evaluation}
\end{table}

\subsection{Human Evaluation and Qualitative Studies}

We recruit 11 participants to score 30 groups of summaries based on their Satisfaction (Satis) and Informativeness (Inform). Each example is scored from 1 to 5 (the higher, the better). The results from Table \ref{tab: summary form} show combining bimodal summary could significantly improve the browsing experience, as TM-Summary and VM-Summary can complement each other. 
To evaluate the performance of UBiSS and pipeline-type baselines, the participants also score 20 randomly selected sets of summarization results in terms of accuracy and consistency, as presented in Table \ref{tab: baseline human evaluation}. UBiSS trained with NeuralNDCG(N)~\cite{Pobrotyn2021NeuralNDCG} outperforms other approaches across all metrics, suggesting a stronger ability to capture highlighted visual-modal information could enhance the quality of summaries. 
The inter-annotator agreement for comparing different forms of summaries is 0.646, and for comparing bimodal summaries generated by UBiSS and baselines is 0.604. Spearman's correlation coefficient is used to compute the agreement. 
More details are in Appendix \ref{sec: human eval}.

Figure ~\ref{fig: case} visualizes bimodal summaries generated by pipeline-type baselines and UBiSS. PGL-Swin suffers from inaccuracies in identifying salient segments, leading to TM-Summary either unrelated to the video (case (a)) or failing to pinpoint the highlight moments (case (b)). While text embeddings assist Swin-PGL in filtering out less relevant visual details, it still encounters challenges with longer videos, as demonstrated in case (b) where the model incorrectly identifies the \textit{hotel room} as the primary focus. In contrast, UBiSS has a superior capability in capturing salient moments, resulting in more accurate summaries.

\section{Conclusion}
\label{sec: conclusion}

We introduce BIDS, a new large-scale biomodal summarization dataset designed to support the BiSSV task.
Based on BIDS, we then propose a unified framework named UBiSS, which harnesses saliency information to enhance the quality of bimodal summaries. A list-wise ranking-based optimization objective~\cite{Pobrotyn2021NeuralNDCG} is employed to help the model learn saliency trends.
Furthermore, we design a novel evaluation metric $NDCG_{MS}$ for joint assessment of bimodal outputs. 
Experiments show that UBiSS excels in capturing salient content, resulting in superior bimodal summarization performance. 
The enriched semantics of bimodal summaries also leads to a more satisfying and informative browsing experience. 
In the future, we will extend our UBiSS to tackle more intricate tasks, such as user-guided multimodal summarization.
\section{Acknowledgments}
This work was partially supported by the the National Natural Science Foundation of China (No. 62072462) and Beijing Natural Science Foundation (No. L233008).

\balance
\bibliographystyle{ACM-Reference-Format}
\bibliography{reference}

\appendix
\clearpage

\section{Appendix}

\subsection{Related Task Comparison}
\label{sec: related task comparison}

\begin{figure*}[b]
    \centering
    \includegraphics[width=0.9\linewidth]{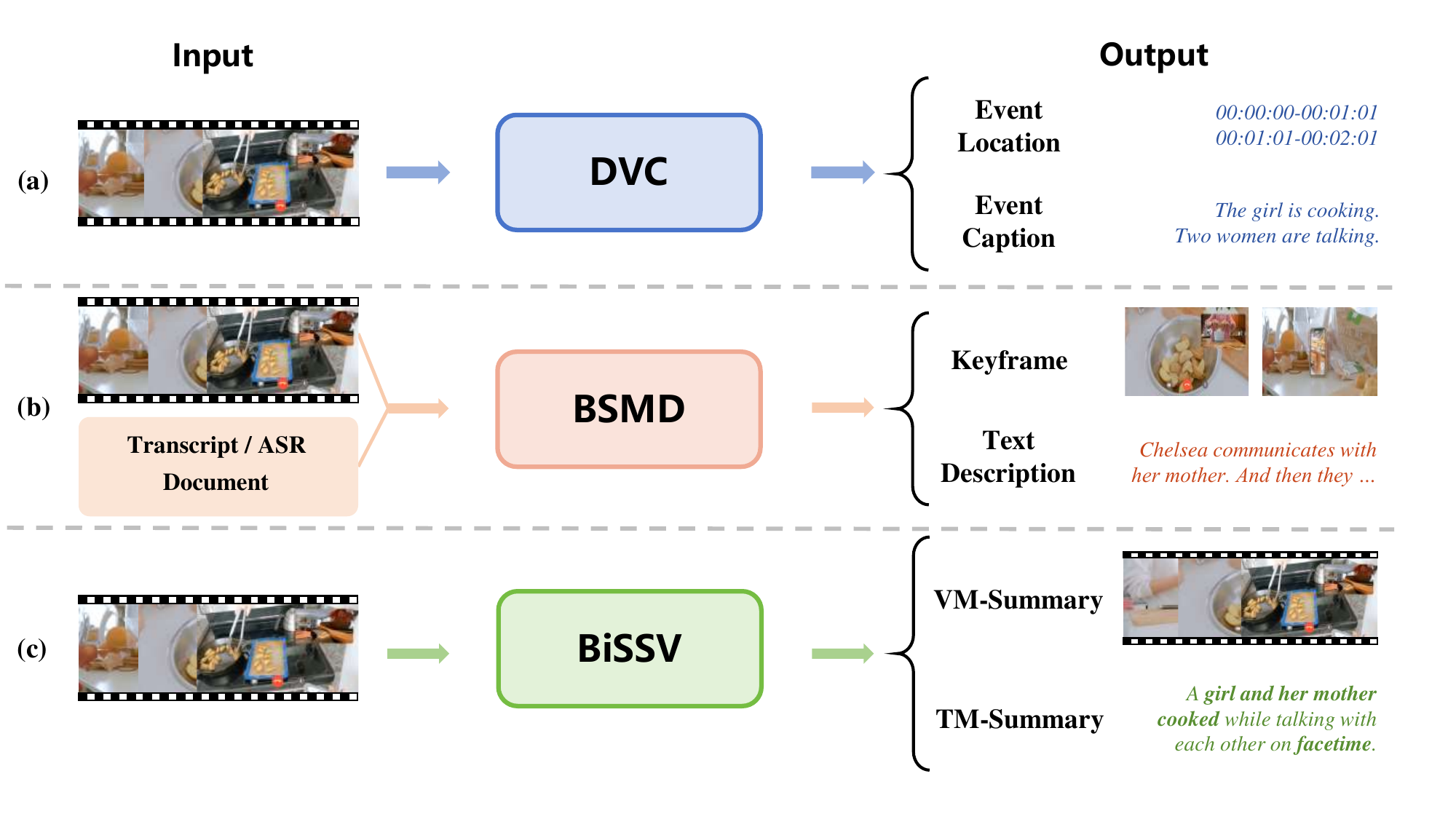}
    \vspace{-8pt}
    \caption{Comparison between BiSSV and related tasks. DVC: Dense Video Captioning. BSMD: Bimodal Summarization of Multimodal Data. BiSSV: Bimodal Semantic Summarization of Videos.}
    \label{fig: task comparison}
\end{figure*}

We could categorize related tasks into four categories based on input/output modality: Textual-Modal Summarization of Videos (TMSV), Visual-Modal Summarization of Videos (VMSV), \textbf{Bi}modal \textbf{S}emantic \textbf{S}ummarization of \textbf{V}ideos (BiSSV), and Bimodal Summarization of Multimodal Data (BSMD). The former two tasks are primarily concerned with unimodal summarization, while BiSSV and BSMD focus on multimodal summarization. Figure \ref{fig: task comparison} illustrates a detailed task comparison.

\textbf{Can unimodal summarization datasets be modified for BiSSV?} One closely related task in unimodal summarization is Dense Video Captioning (DVC)~\cite{krishna2017dense}, which provides text description and grounded segments for the events within the video. DVC could be divided into two sub-tasks: event localization and event captioning~\cite{wang2021end,zhang2022unifying}. The primary divergence between BiSSV and DVC lies in their objectives. The goal of BiSSV is to generate a TM-Summary as a global overview of the entire video instead of chronological captions, and the VM-Summary is a collection of highlighted moments instead of localized events. This difference in focus implies that simply concatenating dense captions from DVC may not produce a suitable TM-Summary, as in the case of the first video-to-video\&text dataset VideoXum~\cite{lin2023videoxum}, which has TM-Summaries that are, on average, 49.9 words in length.

\textbf{Can BSMD datasets be modified for BiSSV?} Recently, the task of multimodal summarization with multimodal output~\cite{zhu2018msmo}, a typical task of BSMD, is proposed to generate a bimodal summary for multimodal inputs. Some BSMD approaches use video and corresponding textual metadata (document~\cite{li2020vmsmo,fu2020multi, tang2023tldw} or transcript~\cite{fu2020multi,he2023align,qiu2023multisum}) as input, yielding a bimodal summary comprising text descriptions and keyframes~\cite{li2020vmsmo,fu2020multi,tang2023tldw,qiu2023multisum} or key segments~\cite{he2023align}. We summarize BSMD datasets with video input in Table \ref{tab: BSMD} for comparison. 

In summary, we discover that BSMD can be more accurately characterized as a combination of unimodal summarization tasks, with a primary focus on information extraction, e.g. direct selection of the most informative sentences from the source text~\cite{he2023align} or detailed description of transcripts~\cite{qiu2023multisum}. Given the time-consuming nature of acquiring auxiliary information in real-world scenarios, BSMD and BiSSV serve distinct application purposes. BSMD is particularly well-suited for video-contained documents, such as news articles, while BiSSV is better suited for various web applications like video browsing, retrieval and recommendation. Furthermore, it's worth mentioning that the majority of visual outputs of BSMD datasets consist of keyframes~\cite{li2020vmsmo,fu2020multi,tang2023tldw,qiu2023multisum}, which lacks smoothness for online browsing compared to short videos.

\begin{table*}[b]
\centering

\begin{tabular}{lccccc}

\toprule
\multirow{2}{*}{} & \multicolumn{2}{c}{\textbf{\textit{Input}}}    & \multicolumn{2}{c}{\textbf{\textit{Output}}} & \multirow{2}{*}{\textbf{Source}} \\
\cmidrule(r){2-3} \cmidrule(r){4-5}
                  & \textbf{Vision} & \textbf{Language}            & \textbf{Vision}       & \textbf{Language}    &                         \\
\midrule
VMSMO~\cite{li2020vmsmo}             & video  & document            & keyframe     & sentence    & news                    \\
MM-AVS~\cite{fu2020multi}            & video  & document/transcript & keyframe    & sentence    & news                    \\
XMSMO~\cite{tang2023tldw}             & video  & document            & keyframe     & sentence    & news                    \\
BLiSS~\cite{he2023align}             & video  & transcript          & segment     & paragraph   & livestream              \\
MultiSum~\cite{qiu2023multisum}          & video  & transcript          & keyframe    & paragraph   & YouTube                 \\
BIDS (ours)             & video  & $\times$            & segment    & sentence    & YouTube  \\
\bottomrule
\end{tabular}
\label{tab: BSMD}
\caption{Comparison between BSMD datasets and BIDS.}
\end{table*}

\subsection{Pseudo-Code for VM-Summary Extraction}
\label{sec: pseudo code}

We provide a pseudo-code for VM-Summary extraction in Algorighm \ref{algorithm: VM extraction}.

\begin{algorithm*}
\caption{VM-Summary Extraction}
\LinesNotNumbered 
\SetKwInOut{Input}{Input}
\SetKwInOut{Output}{Output}

\Input{2s-clips $C_{2s}$, score sequence for 2s-clips $S_{2s}=\{s_i\}$, VM-Summary duration L}
\Output{selected VM-Summary segments $Segment_{VM}$}

$Segments_{all}$ = Merge($C_{2s}$, condition=($s_i==s_i+1$)) \\
$S_{seg}$ = Merge\_Score($S_{2s}$, condition=($s_i==s_i+1$)) \\
For score in Rank($S_{seg}$):\\
\Indp{ 
    $Segment_{score}$ = Get\_Segment($Segments_{all}$, $s_i==score$)\\
    if Duration($Segment_{score}$) + Duration$(Segment_{VM})$ $< L$: \\
    \Indp {
        $Segment_{score}\rightarrow Segment_{VM}$\\
    }
    \Indm
    else:\\
    \Indp{
        For seg in $Segment_{score}$:\\
        \Indp{
            scaled\_length = Duration($seg$) / Duration($Segment_{score}$)\\
            $s_{left}, s_{right}$ = Get\_Score(Left\_Segment(seg), Right\_Segment(seg))\\
            if $s_{left} > score$ and $s_{right} > score$:\\
                \Indp{
                    $Segment_{left}$ = [Get\_Left\_Boundary(seg), Get\_Left\_Boundary(seg)+$scaled\_length/2$] \\
                    $Segment_{right}$ = [Get\_Right\_Boundary(seg)-$scaled\_length/2$, Get\_Right\_Boundary(seg)] \\
                    $Segment_{left}, Segment_{right}\rightarrow Segment_{VM}$ \\
                }
            \Indm
            else if $s_{left} > score$: \\
                \Indp{
                    $Segment_{left}$ = [Get\_Left\_Boundary(seg), Get\_Left\_Boundary(seg)+$scaled\_length$] \\
                    $Segment_{left}\rightarrow Segment_{VM}$\\
                }
            \Indm
            else if $s_{right} > score$: \\
                \Indp{
                    $Segment_{right}$ = [Get\_Right\_Boundary(seg)-$scaled\_length$, Get\_Right\_Boundary(seg)] \\
                    $Segment_{right}\rightarrow Segment_{VM}$\\
                }
            \Indm
            else:\\
                \Indp{
                    $Segment_{mid}$ = [Get\_Mid(seg)-$scaled\_length/2$, Get\_Mid(seg)+$scaled\_length/2$] \\
                    $Segment_{mid}\rightarrow Segment_{VM}$\\
                }
            \Indm
        }
        \Indm
        break\\
    }
    \Indm
}
\Indm
return $Segment_{VM}$
\label{algorithm: VM extraction}
\end{algorithm*}

\subsection{Textual-Modal Summarization Warm-up}
\label{sec: warm up}

\begin{table*}[b]
\begin{tabular}{lcccccc}

\toprule
     & \textbf{CIDEr}          & $\boldsymbol{\tau}$         & $\mathbf{NDCG_{VM}@15\%}$ & $\mathbf{NDCG_{VM}@all}$ & $\mathbf{NDCG_{TM}@15\%}$ & $\mathbf{NDCG_{TM}@all}$ \\
\midrule
N=0  & \underline{38.23}    & 16.14          & 65.99           & 84.74          & 57.09           & \textbf{81.47} \\
N=10 & 36.65          & 16.96          & 66.33           & 84.89          & \textbf{57.39}  & \underline{81.40}    \\
N=20 & 36.91          & \textbf{18.18} & \textbf{67.01}  & \textbf{85.09} & 56.63           & 81.29          \\
N=30 & \textbf{40.87} & 17.58          & \underline{66.58}     & \underline{84.94}    & 56.55           & 81.30          \\
N=40 & 37.52          & \underline{17.61}    & 66.35           & 84.93          & \underline{57.24}     & 81.35          \\
N=50 & 37.74          & 15.70          & 65.20           & 84.45          & 56.88           & 81.35  \\       
\bottomrule
\end{tabular}
\caption{ Comparison of different epochs N for textual-modal summarization warm-up.
}
\label{tab: warm up}
\end{table*}

Results of different epochs for textual-modal summarization warm-up are presented in Table \ref{tab: warm up}, all chosen by CIDEr~\cite{vedantam2015cider}. The model's capability in visual-modal summarization demonstrates an initial improvement followed by a decline as the number of warm-up epochs increases. It is important to note that though the precise number of epochs required for textual-modal summarization warm-up may vary across different machines, a closer alignment between visual-modal and textual-modal summarization consistently yields superior results. Inadequate training or overfitting for each sub-task can lead to a decline in overall performance.

The results presented in Table \ref{tab: warm up} also reveal an issue with the ranking-based optimization objective~\cite{Pobrotyn2021NeuralNDCG}, as discussed in Section \ref{sec: multi-stage}. This issue arises when the model takes shortcuts by assigning extremely low scores to insignificant features. For instance, when considering models trained with N=10 and 20 or 30 warm-up epochs, we observe a consistent increase in $NDCG_{VM}$. However, there is a decline in textual-modal summarization performance, indicating that partially absent features may not be sufficient to generate a global TM-Summary. We address exploring the integration of saliency learning and visual modeling as a promising direction for future research.

\subsection{Human Evaluation Details}
\label{sec: human eval}

\begin{figure*}[b]
    \centering
    \includegraphics[width=0.95\linewidth]{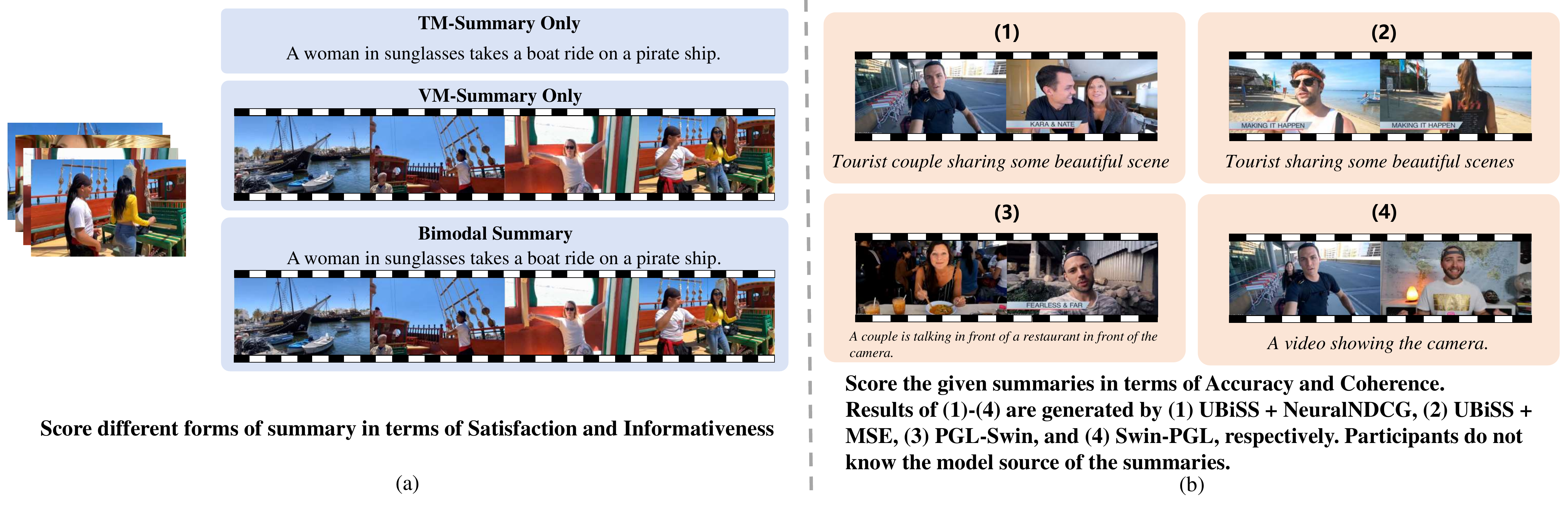}
    \vspace{-8pt}
    \caption{Example cases to be evaluated by participants.}
    \label{fig: human eval}
\end{figure*}

We conduct human evaluation under two settings. To evaluate how bimodal summaries could contribute to Satisfaction (Satis) and Informativeness (Inform) in comparison with unimodal summaries, we randomly select 30 sets of videos and ask participants to score VM-Summary only, TM-Summary only, and bimodal summaries on a scale ranging from 1 to 5. For both metrics, a rating of 1 indicates very dissatisfactory while 5 indicates very satisfactory. For informativeness, a rating of 1 indicates the video content was not summarized adequately while 5 indicates that the video content was perfectly summarized. We present different forms of summaries to participants after shuffling. Figure \ref{fig: human eval} (a) shows an example set of summaries.

Besides automatic evaluation metrics, we also conduct a comparative evaluation between UBiSS and concatenated unimodal summarization baselines. We randomly sample 20 sets of videos. Different summaries generated by UBiSS trained with NeuralNDCG/MSE, PGL-Swin, and Swin-PGL are presented to participants in random order. The participants are asked to rate the accuracy of VM- and TM-Summary, based on how they could capture the highlights (VM-Summary) or present a global overview (TM-Summary), along with the consistency of bimodal summaries. Figure \ref{fig: human eval} (b) offers an illustrative example.

The participants are college students with an educational background in computer science. The average age of the participants is 22, and the gender ratio is 7:4 (males: females). According to DataReportal (due April 2023)\footnote{https://datareportal.com/essential-youtube-stats}, most YouTube users are between 25 and 34, and the gender ratio is approximately 1.195. Therefore, participants' distribution is similar to real-world users' distribution.

\end{document}